%% file: acl_latex.tex
\documentclass[11pt]{article}

\usepackage[final]{acl}

\usepackage{times}
\usepackage{latexsym}

\usepackage[T1]{fontenc}
\usepackage[utf8]{inputenc}

\usepackage{microtype}

\usepackage{inconsolata}

\usepackage{graphicx}

\usepackage{algorithm}
\usepackage{algorithmic}
\usepackage{graphicx}
\usepackage{subcaption}
\usepackage{caption}
\usepackage{booktabs}
\usepackage{multirow}
\usepackage{amsfonts}
\usepackage{amsmath}
\usepackage{enumitem}

\title{Thinking Economically: A Hierarchical Framework for Adaptive-Complexity Reasoning in LLMs}

\author{
  \textbf{Yubo Gao\textsuperscript{1,2}}\thanks{\ These authors contributed equally to this work.},
  \textbf{Haotian Wu\textsuperscript{3}}\footnotemark[1],
  \textbf{Hong Chen\textsuperscript{1,2}},
  \textbf{Junquan Huang\textsuperscript{1}},
\\
  \textbf{Yibo Yan\textsuperscript{1,2}},
  \textbf{Jungang Li\textsuperscript{1}},
  \textbf{Zihao Dongfang\textsuperscript{1}},
  \textbf{Sicheng Tao \textsuperscript{1}},
\\
  \textbf{Puay Siew Tan\textsuperscript{4}},
  \textbf{Jie Zhang\textsuperscript{3}},
  \textbf{Xuming Hu\textsuperscript{1,2\textdagger}}
\\
  \textsuperscript{1}The Hong Kong University of Science and Technology (Guangzhou),\\
  \textsuperscript{2}The Hong Kong University of Science and Technology,\\
  \textsuperscript{3}Nanyang Technological University,\\
  \textsuperscript{4}Singapore Institute of Manufacturing Technology, A*STAR
\\
  \small{
    \textbf{Emails:} \href{mailto:ygao704@connect.hkust-gz.edu.cn}{ygao704@connect.hkust-gz.edu.cn}, \href{mailto:xuminghu@hkust-gz.edu.cn}{xuminghu@hkust-gz.edu.cn}
  }
}

\begin{document}
\maketitle
\begin{abstract}
Chain-of-Thought (CoT) has significantly enhanced LLM reasoning, yet often incurs substantial computational overhead due to ``overthinking'': generating excessively long rationales without commensurate accuracy gains. Existing efficiency methods typically apply uniform compression, which overlooks a critical observation that reasoning complexity is heterogeneous at two distinct granularities: across different problems and within individual reasoning steps. This motivates our principle of \textbf{Thinking Economically}: intelligently allocating computational resources based on intrinsic task and step demands rather than pursuing uniform brevity. We propose Hierarchical Adaptive Budgeter (HAB), a training framework that operationalizes this principle through coarse-to-fine budgeting. At the inter-step level, HAB predicts the optimal reasoning depth for each problem. At the intra-step level, HAB learns step-specific token budgeting signals from PPL-derived step comparisons and an adaptive Pareto optimization objective that captures the local quality-efficiency trade-off, while a Fisher Information-based pruner further provides fine-grained training-time guidance, thereby encouraging the generator to internalize more economical reasoning patterns. Experiments on GSM8K and MATH500 show that HAB not only surpasses standard CoT in accuracy but also reduces token usage, achieving a stronger performance-efficiency trade-off than the compared baselines.
\end{abstract}
\input{Introduction1}
\input{Related_Work}

\input{Methodology}

\input{Experiments}

\section{Conclusions}
We introduce Hierarchical Adaptive Budgeter (HAB), a novel framework that enables LLMs to ``Think Economically" through adaptive reasoning resource allocation. HAB employs a coarse-to-fine budgeting process: it first predicts a coarse reasoning-depth category at the inter-step level and maps it to a step-range instruction, then allocates a dynamic token budget for each step at the intra-step level. The budget allocation is governed by an adaptive Pareto optimization scheme that balances reasoning quality and computational efficiency. Experiments on GSM8K and MATH500 demonstrate that HAB achieves a stronger performance-efficiency trade-off than the compared baselines, improving accuracy while reducing token usage compared to Vanilla CoT.
\section{Limitations}
Our work has several limitations that suggest directions for future research. First, HAB currently focuses on linear CoT structures. Extending the adaptive budgeting paradigm to non-linear reasoning frameworks such as Tree-of-Thoughts~\cite{yao2023tree} and Graph-of-Thought~\cite{besta2024graph} remains an open challenge. Second, while the adaptive Pareto optimization scheme is effective, it incurs modest additional training overhead. Moreover, HAB requires extra data preparation effort to construct the CoT dataset and derive reasoning-depth labels, which currently relies on external LLM-generated candidate chains and subsequent filtering. Developing more lightweight optimization and annotation strategies is therefore a promising direction.

\section{Acknowledgements}
This research is supported by the National Research Foundation, Singapore under its AI Singapore Programme (AISG Award No: AISG3-RP-2022-031), and A*STAR under its MTC Programmatic (Award M23L9b0052). This research is supported in part by the Singapore MOE AcRF Tier 1 funding (RG16/25).

\bibliography{ref}

\appendix

\end{document}

%% file: Introduction1.tex
\section{Introduction}
Large Language Models (LLMs) have demonstrated remarkable reasoning capabilities through Chain-of-Thought (CoT) prompting~\cite{wei2022chain}, which decomposes complex tasks into intermediate steps~\cite{brown2020language,grattafiori2024llama}. To further improve CoT, various extensions have been proposed, including self-consistency sampling~\cite{wang2023self}, Tree-of-Thoughts~\cite{yao2023tree}, Graph-of-Thought~\cite{besta2024graph}, and instruction tuning with reasoning traces~\cite{cai2024system}.
However, these methods introduce significant inefficiencies. Specifically, they often engage in ``overthinking''~\cite{chen2024not,team2025kimi}, generating excessively long reasoning chains without corresponding accuracy gains. Such redundant steps not only increase computational cost but also risk prematurely exhausting the token budget and introducing logical inconsistencies that undermine answer correctness and clarity~\cite{sui2025stop}.

\begin{figure}[htp]
    \centering
    \begin{subfigure}[b]{0.22\textwidth}
        \centering
        \includegraphics[width=\textwidth]{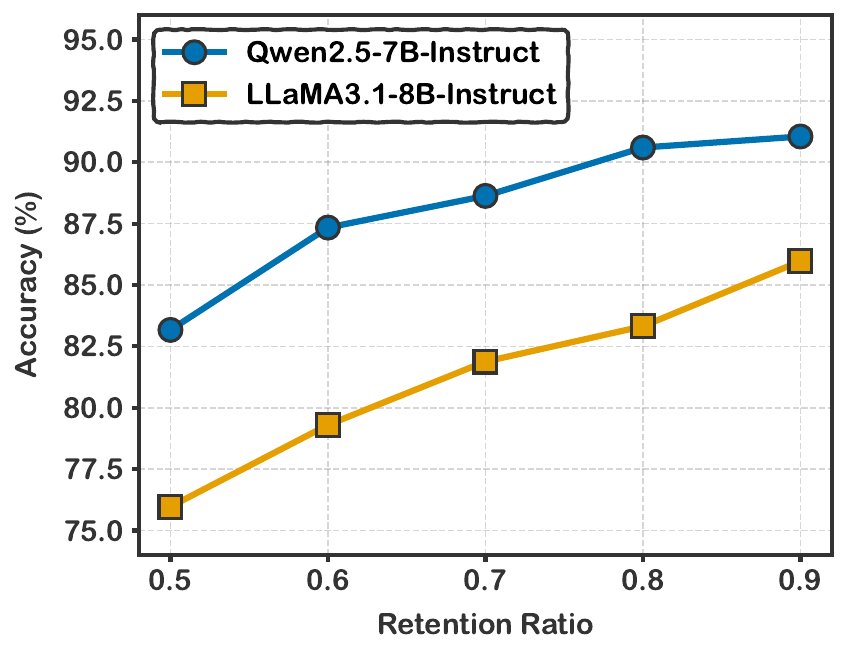}
        \caption{Accuracy}
        \label{fig:accuracy}
    \end{subfigure}
    \hfill
    \begin{subfigure}[b]{0.22\textwidth}
        \centering
        \includegraphics[width=\textwidth]{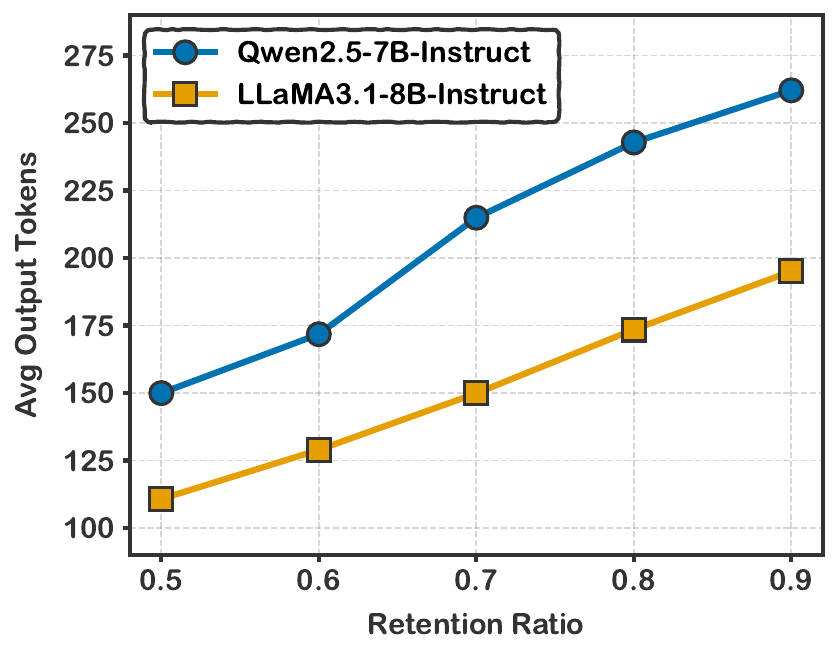}
        \caption{Average Output Tokens}
        \label{fig:tokens}
    \end{subfigure}
    \caption{Comparison of performance of TokenSkip with different LLMs backbones under different retention ratios on the GSM8K dataset.}
    \label{tokenskip}
\end{figure}

A promising direction to address this issue is imposing token constraints or budgets. Existing methods fall into two categories: pre-defined and learning-based constraints. For pre-defined constraints, such as prompt-based length guidance~\cite{xu2025chain,lee2025well,renze2024benefits,chen2024unlocking} or post-processing techniques like TokenSkip~\cite{xia2025tokenskip}, selecting an appropriate retention ratio remains challenging. Our pilot analysis in Figure~\ref{tokenskip} shows that a lower retention ratio reduces token count but causes a sharp drop in accuracy. Therefore, finding a satisfactory balance between efficiency and performance with a static, pre-defined constraint is difficult.

To address this, several methods based on \textbf{learned constraints} have been proposed. These include Reinforcement Learning (RL)~\cite{luo2025o1,yeo2025demystifying} approaches that penalize length, which in practice may be harder to stabilize than direct supervised objectives, and more recent methods like TALE~\cite{han-etal-2025-token} that leverage the LLM's own capabilities to estimate a global budget first, and induce the LLM to reason under such a budget. While a significant step forward, these methods still focus on learning a single, \textit{global} token budget for the entire reasoning chain.

\begin{figure*}[htp]
    \centering
    \begin{subfigure}[b]{0.3\textwidth}
        \centering
        \includegraphics[width=\textwidth]{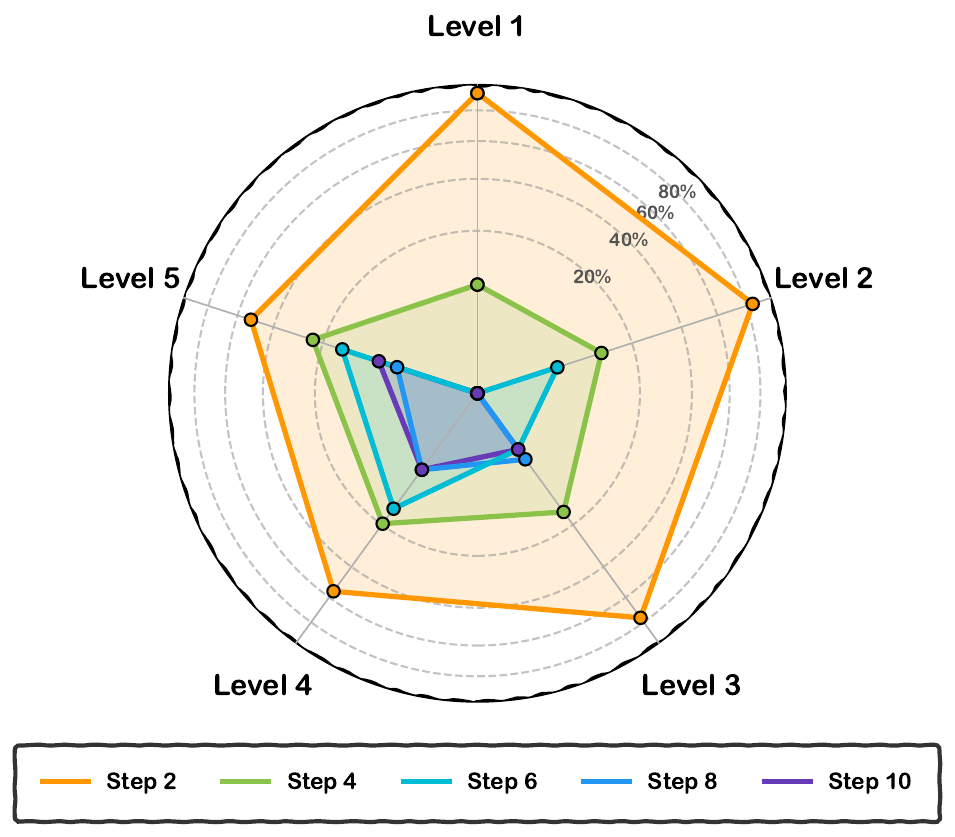}
        \caption{\textbf{Inter-reasoning step}:  Optimal step distribution by difficulty.}
        \label{macro}
    \end{subfigure}
    \hfill
    \begin{subfigure}[b]{0.3\textwidth}
        \centering
        \includegraphics[width=\textwidth]{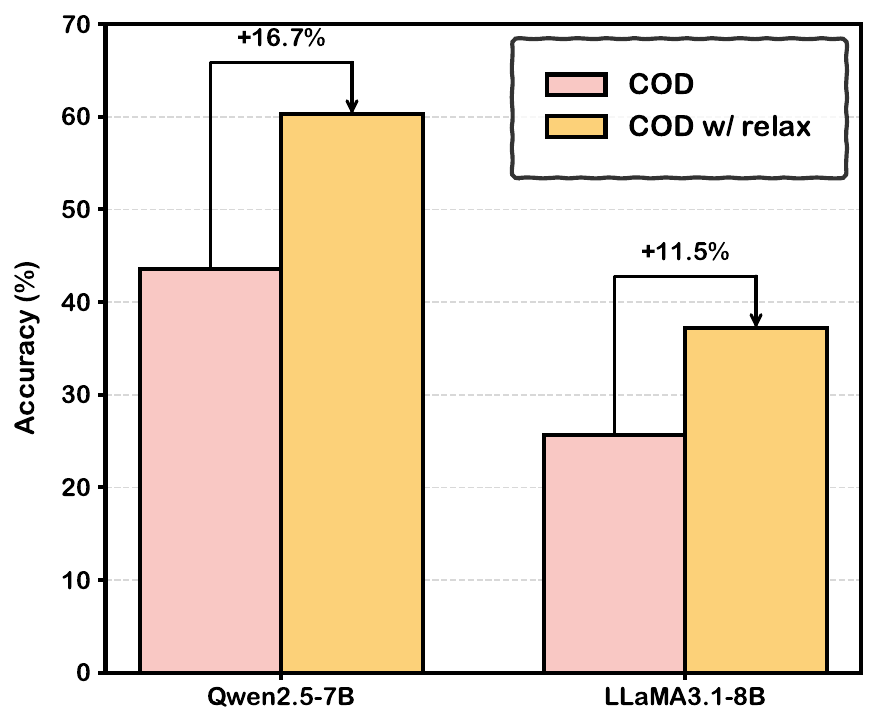}
        \caption{\textbf{Intra-reasoning step}: CoD vs. CoD$_w/_ {\text{relaxation}}$ (Accuracy).}
        \label{math}
    \end{subfigure}
    \hfill
    \begin{subfigure}[b]{0.3\textwidth}
        \centering
        \includegraphics[width=\textwidth]{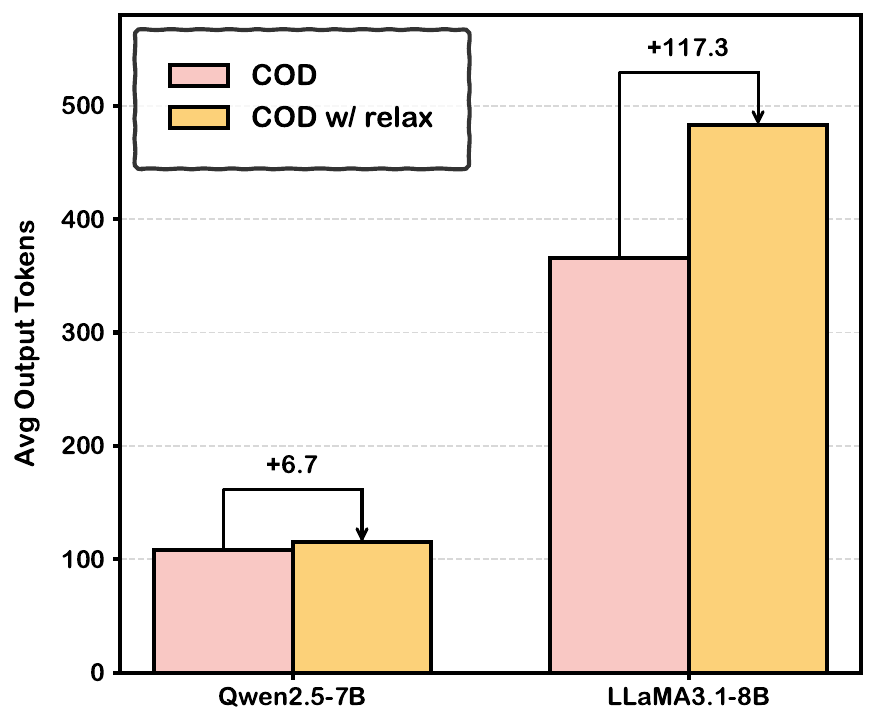}
        \caption{\textbf{Intra-reasoning step}: CoD vs. CoD$_w/_ {\text{relaxation}}$ (Avg output tokens).}
        \label{math}
    \end{subfigure}
    \caption{Exploratory experiments on the necessity of dynamically allocating the number of reasoning steps tailored to problems and allocating token budgets tailored to reasoning steps.}
    \label{motivation}
\end{figure*}

We contend that a global budget, whether pre-defined or learned, is fundamentally limited because it ignores the varying complexity at two granularities (especially the second granularity): \textbf{across different problems} and \textbf{within the reasoning steps of a single problem}. Our analysis on the MATH500 dataset (Figure~\ref{motivation}) provides empirical motivation for hierarchical budgeting. At the inter-step level, shown in Figure~\ref{motivation}(a), simpler problems require fewer reasoning steps, while complex ones need longer chains. At the intra-step level, as shown in Figure~\ref{motivation}(b)(c), our experiments on Chain-of-Draft (CoD) show that allowing the model to allocate more tokens to difficult steps (CoD$_{w/ relaxation}$) leads to significant performance gains. This indicates that different steps have varying token requirements, and a uniform constraint is suboptimal.
These findings motivate our principle of ``Thinking Economically'': instead of uniformly pursuing brevity, we aim to adaptively allocate resources based on hierarchical task demands. The key is to concentrate computation on difficult \textit{problems} and \textit{reasoning steps}, while saving resources on simpler ones.

To realize this vision, we introduce Hierarchical Adaptive Budgeter (HAB), a two-stage training framework for adaptive-complexity reasoning that moves beyond a single global budget toward hierarchical control. HAB first performs inter-step control by predicting a coarse reasoning-depth category for each problem and converting it into a step-range instruction for generation, thereby determining how much reasoning the problem should receive at the global level. It then performs intra-step learning within the generated reasoning process: PPL-derived step comparisons provide supervision on relative step difficulty, an adaptive Pareto optimization objective models the local quality-efficiency trade-off, and a Fisher Information-based pruner supplies additional fine-grained training guidance. Through this coarse-to-fine design, HAB encourages the generator to internalize more economical reasoning behaviors, allocating more computation to harder problems and more demanding steps while avoiding unnecessary verbosity on simpler cases. Our main contributions are:
\begin{itemize}[leftmargin=*]
\item We propose the principle of ``Thinking Economically'' and instantiate it with HAB, a coarse-to-fine hierarchical framework that first performs explicit inter-step reasoning-depth control and then learns finer-grained intra-step efficiency patterns during training.
\item We design an adaptive Pareto optimization scheme to dynamically balance quality and efficiency for each reasoning step.
\item Experiments on GSM8K and MATH500 show that HAB consistently achieves a stronger performance-efficiency trade-off than the compared baselines, improving accuracy while reducing token usage.
\end{itemize}

%% file: Related_Work.tex
\section{Related Work}
\noindent\textbf{Chain-of-Thought Reasoning}. 
Chain-of-Thought (CoT) prompting~\cite{wei2022chain} enhances reasoning in Large Language Models (LLMs) by decomposing solutions of complex problems into intermediate steps. Early works demonstrated its effectiveness in both few-shot~\cite{wei2022chain} and zero-shot~\cite{kojima2022large} settings, but standard CoT often suffers from error propagation due to its linear reasoning path.
To mitigate this limitation, subsequent research has introduced structured reasoning strategies. Least-to-Most prompting~\cite{zhou2023least} addresses compositional generalization by sequentially solving decomposed sub-problems. Tree-of-Thoughts (ToT)~\cite{yao2023tree} enables non-linear exploration through tree-search algorithms with backtracking. Self-Consistency~\cite{wang2023self} improves stability by sampling diverse reasoning trajectories and applying majority voting. Auto-CoT~\cite{zhang2022automatic} further automates exemplar construction via clustering and diversity selection, reducing manual annotation costs.
Beyond inference-time prompting, training-stage methods have been proposed to internalize reasoning capabilities. Symbolic CoT Distillation~\cite{li2023symbolic} transfers reasoning skills from teacher models to smaller students. Self-Taught Reasoner (STaR)~\cite{zelikman2022star} establishes an iterative bootstrapping loop where models improve by training on their own generated rationales.
Recent work addresses practical constraints such as robustness~\cite{liu2024can}, generalization to unseen tasks~\cite{yin2025enhancing}, and inference efficiency~\cite{xu2025softcot}. CoT has also expanded into multimodal domains~\cite{wang2025multimodal} and lightweight adaptation for small-scale models~\cite{zhuang2025unicott}.

\noindent\textbf{CoT Token Constraints.}
Existing methods for controlling token usage fall into two paradigms: pre-defined constraints and learned constraints.
\textbf{Pre-defined Token Constraints.}
These methods fix the efficiency strategy via a static constraint. The simplest form is \emph{prompt-based length guidance}, which requests a token or step budget in the prompt (for example, “use fewer than 50 tokens”) to shorten the rationale~\cite{zhang2025lightthinker,wang2025sampling}. A second line uses \emph{post hoc compression}: the model first produces a long CoT, then removes parts that appear redundant. TokenSkip~\cite{xia2025tokenskip} implements controllable CoT compression by skipping low-importance tokens. C3oT~\cite{kang2025c3ot} trains a compressor and conditions generation on a target compression level to produce shorter CoT. The main limitation is that a single budget or ratio rarely fits all instances. It can truncate hard questions' reasoning or waste tokens on easy ones, which harms the efficiency-accuracy trade-off.

\noindent\textbf{Learning-based Token Constraints.}
To reduce rigidity, learning-based methods optimize the trade-off during training. RL-style approaches incorporate length costs into the objective, so the model learns when to stop or how much to ``think'' for each input, such as DAST~\cite{shen2025dast}, O1-Pruner~\cite{luo2025o1}, and related analyses that study length growth and stabilization in long-CoT training~\cite{yeo2025demystifying}. These methods are adaptive, but they often require careful reward design and may be unstable. TALE~\cite{han-etal-2025-token} moves toward explicit budget control by estimating a global token budget per problem and enforcing it through budget-aware prompting or post-training, which improves the predictability of inference cost.
\textit{Latent reasoning} is a distinct branch within learned constraints. These methods perform intermediate reasoning in hidden states rather than generating explicit tokens/full textual rationale~\cite{hao2024training,cheng2024compressed,shen2025codi,xu2025softcot,zhang2025soft}. This reduces decoding cost substantially but sacrifices interpretability and fine-grained control over computational allocation.

%% file: Methodology.tex
\section{Methodology}
Our method, Hierarchical Adaptive Budgeter (HAB), is founded on the principle of ``Thinking Economically'', moving beyond the prevailing paradigm of uniform chain shortening to instead guide the model to dynamically allocate a computational budget tailored to a task's hierarchically intrinsic demands. HAB operationalizes this principle through a coarse-to-fine hierarchical process. At the \textbf{inter-step} level, it first predicts a coarse reasoning-depth category for each problem and converts it into a step-range instruction for generation. 
Subsequently, at the \textbf{intra-step} level, HAB allocates a specific token budget (retention ratio) to each individual step within the generated reasoning chain, tailored to their intrinsic difficulty. To realize this fine-grained control, HAB first derives step-level difficulty signals from \textbf{PPL-derived step comparisons}, which capture the relative complexity of different reasoning steps. It then learns the corresponding retention signals through an \textbf{adaptive Pareto optimization} objective that models the local quality-efficiency trade-off, encouraging the model to preserve more computation for more demanding steps while compressing simpler ones more aggressively. During training, a \textbf{Fisher Information}-based pruner further provides token-level retention guidance within each step, and the resulting fine-grained compression patterns are internalized into the model parameters for inference. In this way, HAB unifies coarse-grained \textbf{inter-step} reasoning-depth control and fine-grained \textbf{intra-step} budget learning within a hierarchical framework, ultimately encouraging the generator to internalize economical reasoning behaviors.
\begin{figure*}
    \centering
    \includegraphics[width=0.9\linewidth]{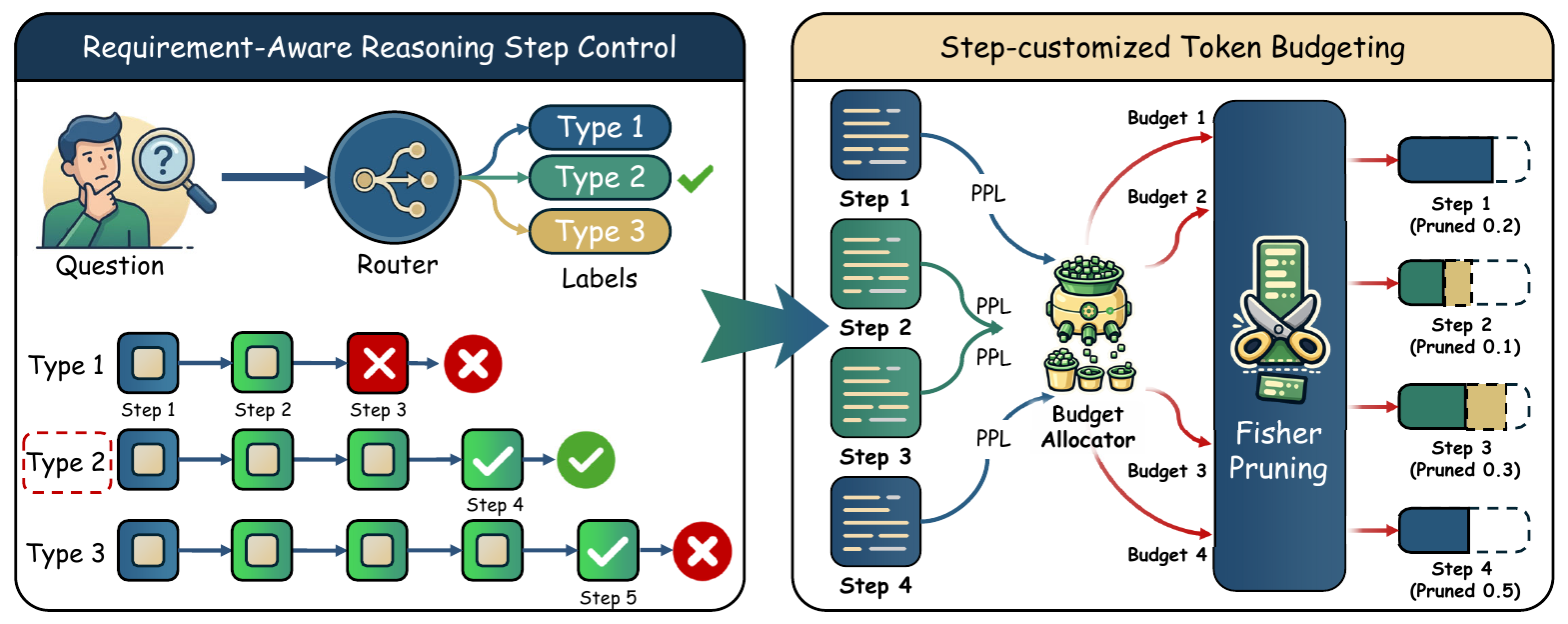}
   \caption{The overall framework of HAB. The Qwen-Max branch is used only during data preparation to construct supervision for reasoning-depth categories; its generated CoT solutions are not directly used as training traces for the downstream model.}
    \label{pipeline}
\end{figure*}
\vspace{-0.2cm}

\subsection{Problem Formulation}
Given an input question $q$ and a LLM $\mathcal{M}$ with parameters $\theta$, HAB performs hierarchical control at two levels. At the \textbf{inter-step} level, the model predicts a coarse reasoning-depth category
$C_{\mathrm{opt}} = f_{\mathrm{macro}}(q;\theta)$,
where $C_{\mathrm{opt}}$ denotes the preferred reasoning-depth category (e.g., short, medium, or long). This predicted category is then mapped to a step-range instruction for generation. At the \textbf{intra-step} level, HAB learns step-level retention signals over training-time reasoning steps $S=\{s_1,s_2,\ldots,s_N\}$, where each step $s_j$ consists of $L_j$ tokens. These retention signals are optimized during Stage-2 training to capture fine-grained quality-efficiency trade-offs and are subsequently internalized into the generator parameters for inference.

\subsection{Inter-step: Requirement-Aware Reasoning Step Control}
\label{sec:methodology}

The first stage of HAB addresses the coarse-grained task of determining the optimal length of the reasoning path. The objective is to train the model to predict the most resource-efficient reasoning depth category, $C_{opt}$, which corresponds to the optimal number of steps, $N_{opt}$, required to solve a given problem correctly.
To achieve this, we construct a high-quality dataset, $\mathcal{D}_{SCP}$, that is utilized to train the model for reasoning depth prediction. For each question $q_i$ from benchmarks such as MATH500 and GSM8K, we first use a powerful LLM (e.g., Qwen-Max) to generate a portfolio of CoT solutions with a varying number of steps.
To ensure that the step count reflects genuine reasoning granularity rather than superficial formatting, we impose a constraint on the average word count per step during generation. After verifying the correctness of each generated chain against the ground-truth answer, we select the chain with the fewest reasoning steps as the optimal reasoning chain (\emph{Questions for which no correct reasoning chain was generated are excluded}). The step count from this chain is defined as the optimal number of steps, $N_{opt,i}$. 
Finally, we apply a series of data cleaning procedures, including standardizing step delimiters, removing redundant formatting, and normalizing mathematical expressions, to ensure dataset quality.

As shown in Figure~\ref{motivation}(a), the distribution of optimal step lengths is heavily skewed, causing severe class imbalance. To address this, we group step counts into three balanced categories: short, medium, and long. The thresholds are dataset-specific: for \textbf{MATH}, we define 1-2 steps as \emph{short}, 3-4 as \emph{medium}, and 5+ as \emph{long}; for \textbf{GSM8K}, the thresholds are 1, 2, and 3+ steps respectively. This bucketing transforms sparse step prediction into a balanced classification problem, enabling more robust learning.

During training, we employ a LoRA-tuned LLM-based \textbf{Router} to classify the reasoning depth of a given problem. It outputs a probability distribution $p_i$ over our three pre-defined categories: {\emph{short}, \emph{medium}, \emph{long}}. We optimize this classification task using a standard Cross-Entropy (CE) loss, $\mathcal{L}_{SCP}$. Let $y_i$ be the one-hot vector representing the ground-truth category for a given problem. The loss is computed over each training batch $\mathcal{B}$ as:
\begin{equation}
\mathcal{L}_{\mathrm{SCP}} = -\frac{1}{|\mathcal{B}|} \sum_{i \in \mathcal{B}} \sum_{k=1}^{3} y_{i,k} \log(p_{i,k}).
\end{equation}
At inference time, the Router first predicts the most likely reasoning-depth category (e.g., `\emph{medium}'). This prediction is then mapped to its corresponding step-count range under our bucketing rules and incorporated into the prompt as an explicit instruction, which guides the downstream LLM to reason within the predicted step range. This provides an elastic budget for the number of steps rather than an overly rigid one. For instance, if the Router predicts the `\emph{medium}' category for a MATH problem, the prompt becomes: ``Let's solve the problem step-by-step with 3 to 4 steps.'' Notably, the role of Qwen-Max is only to construct the supervision signal for reasoning-depth categories during data preparation; the CoT solutions generated by Qwen-Max are not directly used as reasoning traces during downstream generation.

\subsection{Intra-step: Step-customized Token Budgeting}
Once the inter-step plan is established, this stage learns a fine-grained, dynamic token budgeting mechanism over the reasoning process under the predicted step-range instruction. Our approach involves a two-phase process: (1) we first learn a robust indicator of each step's intrinsic complexity, and (2) we then use an adaptive Pareto optimization scheme to translate this complexity into a token budget (retention ratio) that balances performance and efficiency, with a Fisher Information-based pruner further providing token-level retention guidance during training.

\subsubsection{Step Complexity Indicator}

\noindent\textit{Complexity Metric Definition.}
Effective budget allocation requires accurate estimation of each step's intrinsic complexity. We develop a learnable complexity estimator that captures the computational demands of individual reasoning steps. Inspired by prior work~\cite{cui2025stepwise}, we adopt perplexity (PPL) as a practical complexity proxy, as it correlates with the relative difficulty of different reasoning steps. For step $s_{i,j}$ (the $j$-th step of sample $i$), we compute:
\begin{equation}
\tiny
\mathrm{PPL}(s_{i,j}) = \exp
\left(
-\frac{1}{L_{i,j}}
\sum_{k=1}^{L_{i,j}}
\log P(t_{i,j,k}\mid \mathrm{context}_{<(i,j,k)}; \theta)
\right),
\end{equation}
where $L_{i,j}$ is the token length of step $s_{i,j}$, $t_{i,j,k}$ is the $k$-th token in step $s_{i,j}$, and $\mathrm{context}_{<(i,j,k)}$ includes all preceding tokens. In implementation, step-wise PPL is computed during Stage-2 data preparation and used only as a training-time supervision signal.

\noindent\textbf{Learning Complexity Indicator.}
We train a difficulty prediction head $h_{\mathrm{diff}}:\mathbb{R}^{d}\rightarrow\mathbb{R}$ to estimate step complexity from the last hidden states. Direct regression on raw PPL values often suffers from high variance and poor generalization. Instead, to capture relative difficulty differences between reasoning steps, we adopt a ranking-based objective. For each sample $i$ with $N_i$ steps, we consider all $\frac{N_i(N_i-1)}{2}$ unique pairs of steps $(s_{i,j}, s_{i,k})$. To further refine the learning signal, we treat pairs with smaller PPL differences as harder to distinguish and therefore assign them larger weights. Specifically, we define
\begin{equation}
w_{jk}=
\frac{1}{|\mathrm{PPL}(s_{i,j})-\mathrm{PPL}(s_{i,k})|+\epsilon},
\end{equation}
where $\epsilon$ is a small smoothing constant. These weights are then normalized within each sample using a softmax function to produce $w'_{jk}$. The final ranking loss encourages the predicted score of the harder step, $d_{\mathrm{hard}}$, to be greater than that of the easier one, $d_{\mathrm{easy}}$, by at least a margin $m$:
\begin{equation}
\small
\mathcal{L}_{\mathrm{pair}}=
\frac{1}{|\mathcal{B}|}
\sum_{i\in\mathcal{B}}
\sum_{(j,k)}
w'_{jk}\cdot
\max\!\left(0,\; m-(d_{\mathrm{hard}}-d_{\mathrm{easy}})\right),
\end{equation}
where $d_{i,j}=h_{\mathrm{diff}}(\mathrm{StepEncoder}(s_{i,j}))$ is the predicted difficulty score, and the harder/easier ordering is determined by the relative PPL ranking of the compared steps.

The resulting latent score $d_{i,j}$ is an unbounded value. To map this score to a normalized token retention ratio $r_{i,j}\in[0,1]$, we apply a learnable affine transformation followed by a sigmoid function (Budget Allocator):
\begin{equation}
r_{i,j} = \sigma(w\cdot d_{i,j}+b).
\end{equation}
This provides a simple and effective trainable mapping from a raw difficulty score to an actionable token budget, and in implementation serves as the budget allocator during Stage-2 training.

\subsubsection{Adaptive Pareto Optimization for Budget Allocation}
The token budget for each step, represented by a retention ratio $r_{i,j} \in [0, 1]$, is learned by balancing two conflicting objectives: reasoning quality ($\mathcal{L}_{\text{qual},i}$) and computational efficiency ($\mathcal{L}_{\text{eff},i}$). Intuitively, complex steps require more tokens to preserve quality, while simple steps can be compressed more aggressively with minimal quality loss. Traditional methods use fixed trade-off weights, which cannot adapt to varying step complexities. Pareto optimization provides a principled perspective for such multi-objective problems by characterizing the trade-off between competing objectives~\cite{sener2018multi, lin2019pareto, pimentel2020pareto, zhou2024autopeft}. Inspired by this perspective, we design an \textit{adaptive Pareto optimization} scheme that dynamically adjusts the quality-efficiency balance according to local step difficulty.

First, we define the two conflicting losses. \textbf{Quality} is measured by the standard language modeling loss (negative log-likelihood) under pruning, while \textbf{efficiency} is the average token retention ratio:
\begin{equation}
\small
\begin{split}
    \mathcal{L}_{\text{qual}, i} &= -\sum_{j=1}^{N_i} \sum_{k=1}^{L_{i,j}} \log P(t_{i,j,k} \mid S'_{\text{pruned}, i, <(j,k)}) \\
    \mathcal{L}_{\text{eff}, i} &= \frac{1}{N_i} \sum_{j=1}^{N_i} r_{i,j}.
\end{split}
\end{equation}

To find the optimal trade-off, we introduce a \textbf{Pareto Curvature Probe}. The core idea is to estimate the marginal cost of efficiency: ``\emph{If we tighten our budget by a tiny amount $\delta$, how much quality do we lose?}'' We approximate this using a lightweight perturbation-based probe to compute the local slope ($\phi_i$) of the Pareto frontier:
\begin{equation}
\label{eq:slope}
\phi_i = \frac{\mathcal{L}'_{\text{qual}, i} - \mathcal{L}_{\text{qual}, i}}{\delta},
\end{equation}
where $\mathcal{L}'_{\text{qual}, i}$ is the quality loss under a more aggressive budget $r'_{i,j}=\max(r_{i,j}-\delta, 0)$. This slope quantifies the marginal cost of efficiency. A high, positive slope indicates that we are in a ``steep'' region of the trade-off curve, where even small efficiency gains are very costly to quality. A low slope indicates that we are in a ``flat'' region, where efficiency can be improved at relatively low quality cost. We use this signal to dynamically set the loss weights ($\alpha_\mathrm{qual}, \alpha_\mathrm{eff}$).
\begin{equation}
    \begin{array}{c}
\alpha_{\mathrm{eff}, i}=\sigma\left(-k \cdot \phi_{i}+b^{\prime}\right) \\
\alpha_{\mathrm{qual}, i}=1-\alpha_{\mathrm{eff}, i}.
\end{array}
\end{equation}
This sigmoid function acts as a smooth switch. When the slope is high, the input to the sigmoid becomes a large negative value, driving $\alpha_\mathrm{eff}$ towards 0. This forces the optimizer to prioritize quality. Conversely, when the slope is low, $\alpha_\mathrm{eff}$ increases, encouraging the model to pursue higher efficiency. This leads to the final Pareto-balanced loss:
\begin{equation}
\label{eq:pareto_loss}
\mathcal{L}_{\text{pareto}, i} = \alpha_{\text{qual}, i} \cdot \mathcal{L}_{\text{qual}, i} + \alpha_{\text{eff}, i} \cdot \mathcal{L}_{\text{eff}, i}.
\end{equation}

\subsubsection{Pruning Execution via Fisher Information}
During Stage 2 training, once the retention ratio $r_{i,j}$ for step $s_{i,j}$ is determined, we perform token-level pruning within that step. Specifically, we retain the tokens that are most critical to the model's reasoning process. The importance of each token $t_{i,j,k}$ is approximated by the Fisher Information~\cite{brunel1998mutual,ohno2024adaptive}, which we estimate using the squared norm of the gradient of the quality loss with respect to the token embedding $\mathbf{e}(t_{i,j,k})$~\cite{liu2021group}:
\begin{equation}
\label{eq:fisher}
I(t_{i,j,k}) \approx \left\| \nabla_{\mathbf{e}(t_{i,j,k})} \mathcal{L}_{\text{qual},i} \right\|^2.
\end{equation}
For each step $s_{i,j}$, we compute token importance scores, rank all tokens within the step, and retain the top $k_{i,j} = \lfloor L_{i,j} \cdot r_{i,j} \rfloor$ tokens to form the pruned step $s'_{i,j}$. This Fisher-guided pruning is used only during training to provide fine-grained token-level compression supervision; during inference, the learned compression behavior is internalized into the model parameters rather than executed through explicit pruning.

\subsection{Staged Training Strategy}
We train HAB via two-stage sequential fine-tuning with LoRA to reduce task interference between inter-step planning (Stage 1) and intra-step budgeting (Stage 2). Stage 1 learns a global skill, namely classifying the appropriate reasoning-depth category for the entire problem, and is optimized with the cross-entropy loss $\mathcal{L}_{\mathrm{SCP}}$. Stage 2 then learns a local skill: estimating step-level retention ratios and performing Fisher-guided token pruning within each step to support adaptive budgeting. Training both stages jointly with a unified objective could introduce destructive gradient interference, since global reasoning-depth classification and local token-level compression follow different learning dynamics. Therefore, we first adapt the model to predict reasoning depth, and then continue fine-tuning for intra-step adaptive budgeting using the combined objective $\lambda_1 \mathcal{L}_{\mathrm{pair}} + \lambda_2 \mathcal{L}_{\mathrm{pareto}}$.

%% file: Experiments.tex
\section{Experiments}
\subsection{Baselines and Evaluation Metrics}
\label{sec:baselines}
We compare HAB against six representative methods spanning diverse strategies.
(1) \textbf{Vanilla CoT (Zero-shot)}~\cite{wei2022chain}: the standard zero-shot CoT prompting baseline, which generates unconstrained reasoning paths and serves as our primary reference point for both performance and efficiency.
\textbf{Pre-defined constraint methods}: (2) \textbf{TokenSkip}~\cite{xia2025tokenskip}: prunes a fixed ratio of tokens from CoT chains and fine-tunes on compressed data. (3) \textbf{Chain of Draft (CoD)}~\cite{xu2025chain}: instructs the model to generate minimalistic drafts for each step. (4) \textbf{Skeleton-of-Thought (SoT)}~\cite{ning2024skeleton}: first generates an answer skeleton, then expands each point in parallel. (5) \textbf{Sketch-of-Thought}~\cite{aytes2025sketch}: uses a router to select a reasoning paradigm from a predefined library.
\textbf{Learning-based constraint methods:} (6) \textbf{O1-Pruner}~\cite{luo2025o1}: an RL-based method using length-harmonizing reward within PPO to shorten reasoning paths.
Following prior work~\cite{han-etal-2025-token}, we use two metrics: \emph{Accuracy}, the percentage of correctly solved problems via exact match, and \emph{Average output tokens}, the mean number of generated tokens per problem.

\subsection{Implement Details and Datasets}
Experiments are conducted on dual NVIDIA A800 GPUs (80GB each). We apply LoRA~\cite{hu2022lora} fine-tuning to Qwen2.5-7B-Instruct\footnote{https://huggingface.co/Qwen/Qwen2.5-7B-Instruct} and Llama3.1-8B-Instruct\footnote{https://huggingface.co/meta-llama/Llama-3.1-8B-Instruct}, targeting attention projection layers ($q_{proj}$, $k_{proj}$, $v_{proj}$, $o_{proj}$) with rank 16, alpha 32, and dropout 0.1. For Stage 1 (inter-step), we train for 5 epochs with learning rate of $3 \times 10^{-5}$, batch size 4, gradient accumulation steps 2, and weighted cross-entropy loss to address class imbalance. For Stage 2 (intra-step), we train for 3 epochs with learning rate $2 \times 10^{-5}$, $\lambda_1$=0.03, $\lambda_2$=0.01, $m$=0.5, and $\delta$=0.05. All models use AdamW with weight decay 0.01, gradient clipping 1.0, and maximum sequence length 512.

\textbf{Datasets.} Following prior work~\cite{munkhbat2025self,xia2025tokenskip}, we evaluate on two mathematical reasoning benchmarks. (1) \textbf{MATH500}~\cite{hendrycks2021measuring}: 500 high-school competition problems, split into 233/77/78 for train/val/test after excluding instances without a verifiable correct reasoning chain in Qwen-Max (see Section~\ref{sec:methodology}). (2) \textbf{GSM8K}~\cite{cobbe2021training}: grade-school math word problems; we randomly sample 2{,}000 instances and, after applying the same filtering procedure, remaining examples are split into 1007/434/462 for train/val/test.

\subsection{Results and Discussions}
\paragraph{Comparison with Baselines.}
Our experimental evaluation, summarized in Table 1 for the GSM8K dataset and Table 2 for the MATH500 dataset, provides a comprehensive comparison of HAB against a suite of SOTA baselines.

\begin{table}[htp]
\centering
\caption{Performance comparison on GSM8K dataset.}
\label{tab:gsm8k}
\resizebox{\columnwidth}{!}{%
\begin{tabular}{lcccc}
\toprule
\multirow{2}{*}{Method} & \multicolumn{2}{c}{Qwen2.5-7B-Instruct} & \multicolumn{2}{c}{Llama3.1-8B-Instruct} \\
\cmidrule(lr){2-3} \cmidrule(lr){4-5}
& Acc (\%) & Tokens & Acc (\%) & Tokens \\
\midrule
Vanilla CoT (Zero-Shot) & \underline{93.94} & 283.0 & \underline{83.16} & 244.5 \\
\midrule
CoD & 74.24 & \textbf{44.2} & 64.29 & \underline{124.3} \\
Skeleton-of-Thought & 71.65 & 105.1 & 61.04 & 290.4 \\
TokenSkip & 90.04 & 165.9 & 81.17 & \underline{124.3} \\
Sketch-of-Thought & 73.81 & \underline{97.3} & 62.55 & \textbf{112.8} \\
O1-Pruner & 84.63 & 198.7 & 79.22 & 205.4 \\
\midrule
HAB (Ours) & \textbf{95.24} & 211.1 & \textbf{87.23} & 232.2 \\
\bottomrule
\end{tabular}%
}
\end{table}
\vspace{-0.2cm}
\begin{table}[htp]
\centering
\caption{Performance comparison on MATH500 dataset.}
\label{tab:math500}
\resizebox{\columnwidth}{!}{%
\begin{tabular}{lcccc}
\toprule
\multirow{2}{*}{Method} & \multicolumn{2}{c}{Qwen2.5-7B-Instruct} & \multicolumn{2}{c}{Llama3.1-8B-Instruct} \\
\cmidrule(lr){2-3} \cmidrule(lr){4-5}
& Acc (\%) & Tokens & Acc (\%) & Tokens \\
\midrule
Vanilla CoT (Zero-Shot) & \underline{79.49} & 482.2 & 46.15 & 572.9 \\
\midrule
CoD & 43.59 & \textbf{108.3} & 25.64 & \underline{365.5} \\
Skeleton-of-Thought & 47.44 & \underline{135.3} & 39.74 & 529.7 \\
TokenSkip & 75.64 & 354.6 & \underline{48.72} & \textbf{282.2} \\
Sketch-of-Thought & 52.56 & 237.3 & 35.9 & 418.5 \\
O1-Pruner & 69.23 & 350.8 & 47.44 & 381.7 \\
\midrule
HAB (Ours) & \textbf{82.05} & 327.3 & \textbf{51.28} & 424.8 \\
\bottomrule
\end{tabular}%
}
\end{table}
\textbf{First}, HAB achieves superior performance-efficiency trade-offs on both datasets. On GSM8K with Qwen2.5-7B-Instruct, HAB improves accuracy over Vanilla CoT while reducing token usage. This simultaneous improvement in both metrics distinguishes HAB from all baselines, which invariably sacrifice one for the other.
\textbf{Second}, pre-defined constraint methods suffer severe accuracy degradation. For example, CoD shows substantial drops on both GSM8K and MATH500, sacrificing a large amount of accuracy in exchange for aggressive token reduction, highlighting the limitations of uniform compression approaches.
\textbf{Third}, O1-Pruner, representing RL-based approaches, underperforms HAB despite comparable token usage. This indicates that global budget optimization, even when learned through RL, cannot match HAB's hierarchical coarse-to-fine budgeting control in our setting.
\textbf{Finally}, cross-model evaluation with Llama3.1-8B-Instruct shows that HAB remains effective across backbones. HAB consistently outperforms all baselines on both GSM8K and MATH500. Notably, HAB uses more tokens on MATH500 than on GSM8K, demonstrating its ability to adaptively allocate more computation to harder problems.

\subsubsection{Broader Adaptability for HAB}
\begin{figure}[htp]
    \centering
    \includegraphics[width=0.9\linewidth]{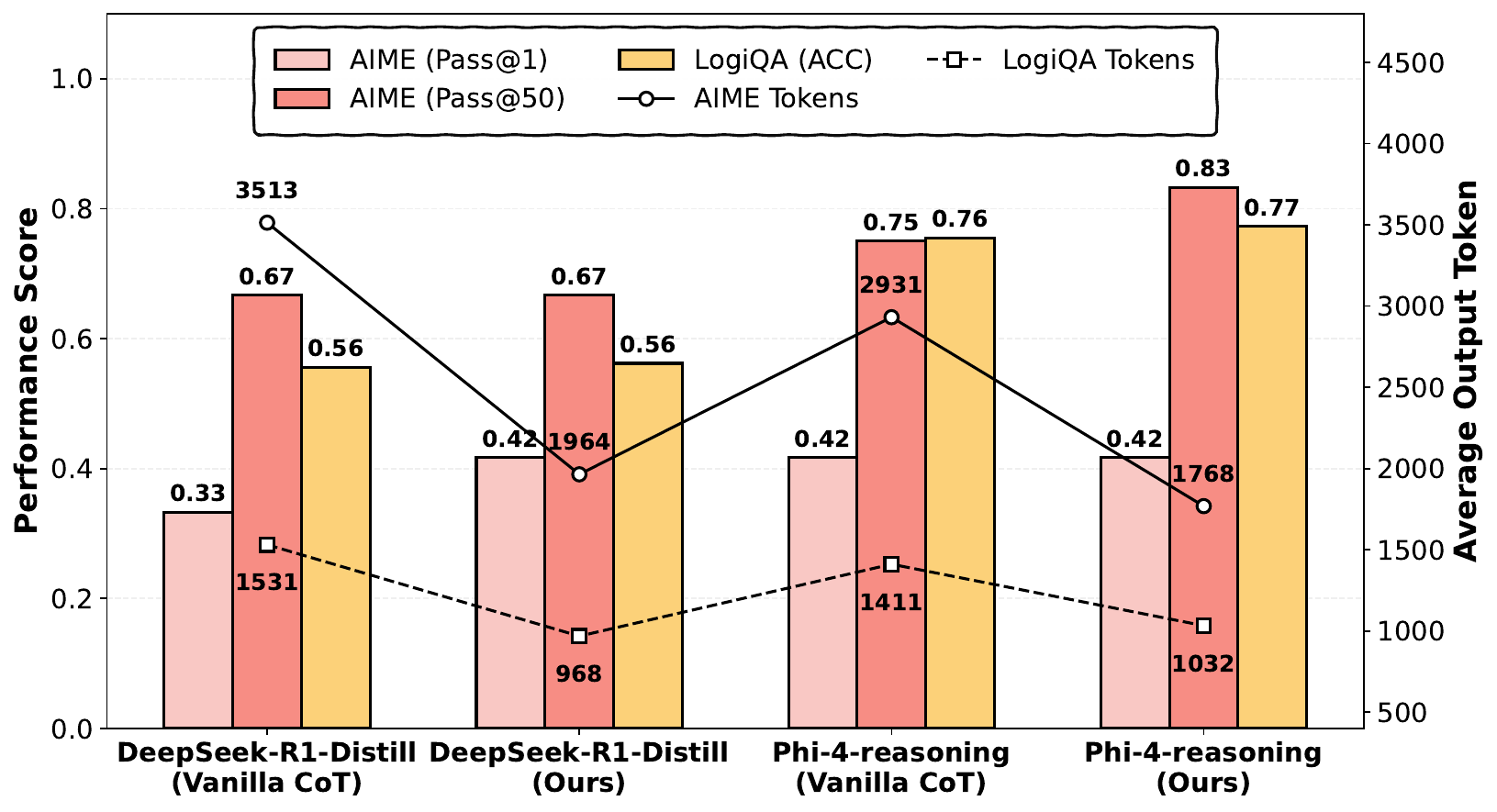}
    \caption{Broader adaptability of HAB across datasets and reasoning backbones.}
    \label{4.3.1}
\end{figure}
To further demonstrate the robustness of HAB, we extend our evaluation to specialized reasoning-oriented models and high-difficulty benchmarks. The objective is to verify whether the "Thinking Economically" principle holds value for models already optimized for complex reasoning, which are still prone to ``overthinking'' and generating redundant computational overhead.
We incorporate DeepSeek-R1-Distill-Qwen-7B\footnote{https://huggingface.co/deepseek-ai/DeepSeek-R1-Distill-Qwen-7B} and Phi-4-reasoning\footnote{https://huggingface.co/microsoft/Phi-4-reasoning} as backbones, testing them on the AIME (mixture of AIME 2025\footnote{https://huggingface.co/datasets/opencompass/AIME2025} and 2024\footnote{https://huggingface.co/datasets/Maxwell-Jia/AIME\_2024}, 15/5/12 for train/val/test after filter) and LogiQA~\cite{liu2020logiqa} datasets (1172/381/331 for train/val/test after filter). As shown in Figure~\ref{4.3.1}, HAB successfully maintains or even improves the accuracy of these advanced models while significantly reducing token usage. This demonstrates that even for models with strong inherent reasoning capabilities, HAB's hierarchical budgeting effectively identifies and prunes logical redundancies. These results confirm that our framework is a versatile paradigm applicable to the next generation of reasoning-heavy models and highly complex logical tasks.

\subsubsection{Ablation Studies}
The ablation study demonstrates that HAB achieves optimal efficiency-performance balance through adaptive step prediction, a capability that fixed-step configurations fundamentally lack. While certain fixed configurations occasionally match or marginally exceed HAB's accuracy, they invariably do so at the expense of computational efficiency, validating our core principle of thinking economically.
\begin{table}[htp]
\centering
\caption{Impact of Fix Length parameter on our HAB's performance.}
\label{tab:fix_length}
\resizebox{\columnwidth}{!}{%
\begin{tabular}{lcccc}
\toprule
\multicolumn{5}{c}{\textbf{Fix Length\_Type = 0}} \\
\midrule
Model & \multicolumn{2}{c}{GSM8K} & \multicolumn{2}{c}{MATH500} \\
\cmidrule(lr){2-3} \cmidrule(lr){4-5}
& Acc (\%) & Tokens & Acc (\%) & Tokens \\
\midrule
Qwen2.5-7B-Instruct & 95.02 & 172.6 & 79.21 & 309.5 \\
Llama3.1-8B-Instruct & 84.42 & 159.6 & 48.15 & 313.8 \\
\midrule
\multicolumn{5}{c}{\textbf{Fix Length\_Type = 1}} \\
\midrule
Model & \multicolumn{2}{c}{GSM8K} & \multicolumn{2}{c}{MATH500} \\
\cmidrule(lr){2-3} \cmidrule(lr){4-5}
& Acc (\%) & Tokens & Acc (\%) & Tokens \\
\midrule
Qwen2.5-7B-Instruct & 95.89 & 310.9 & 81.77 & 496.6 \\
Llama3.1-8B-Instruct & 87.01 & 245.1 & 50.72 & 478.2 \\
\midrule
\multicolumn{5}{c}{\textbf{Fix Length\_Type = 2}} \\
\midrule
Model & \multicolumn{2}{c}{GSM8K} & \multicolumn{2}{c}{MATH500} \\
\cmidrule(lr){2-3} \cmidrule(lr){4-5}
& Acc (\%) & Tokens & Acc (\%) & Tokens \\
\midrule
Qwen2.5-7B-Instruct & 95.45 & 316.3 & 80.49 & 514.7 \\
Llama3.1-8B-Instruct & 89.18 & 290.8 & 49.44 & 563.3 \\
\bottomrule
\end{tabular}%
}
\end{table}
On GSM8K with Qwen2.5-7B-Instruct, the 1-step configuration achieves slightly higher accuracy than HAB. However, this marginal gain comes at a substantially higher token cost, indicating a much weaker efficiency-performance trade-off. Similarly, on MATH500 with Llama3.1-8B-Instruct, the 2-step configuration underperforms HAB while also consuming more tokens. These results highlight the inefficiency of static policies that cannot differentiate computational investment based on problem complexity.

Furthermore, the experimental results reveal a critical insight: more reasoning steps do not monotonically improve performance. With Qwen2.5-7B-Instruct on MATH500, accuracy decreases when moving from 1 step to 2 steps, despite increased token usage. This non-monotonic relationship suggests that excessive reasoning steps can introduce errors or logical inconsistencies, further illustrating the overthinking phenomenon. The model's best performance therefore requires not simply more steps, but an appropriate number of steps for each specific problem.
HAB's adaptive mechanism naturally captures this complexity by learning to allocate reasoning budget based on hierarchical task requirements rather than applying uniform treatments. The framework achieves consistently strong performance across both datasets while maintaining computational efficiency, avoiding both the under-reasoning that can hurt complex problems and the over-reasoning that wastes resources on simpler ones.